\documentclass[sigconf,nonacm]{acmart}
\usepackage{lipsum}  
\usepackage{svg}

\usepackage{balance} %

\usepackage{algorithm}
\usepackage{algorithmic}
\usepackage{tikz}
\usetikzlibrary{cd}

\usepackage{color}

\usepackage{graphicx}
\usepackage{subcaption}

\title{Adaptive Preference Aggregation}

\author{Benjamin Heymann}
\affiliation{
  \institution{Criteo AI Lab}
  \country{France}}
\email{b.heymann@criteo.com}

\begin{abstract}
AI alignment, the challenge of ensuring AI systems act in accordance with human values, has emerged as a critical problem in the development of systems such as foundation models and recommender systems. Still, the current dominant approach, reinforcement learning with human feedback (RLHF) faces known theoretical limitations in aggregating  diverse human preferences.  
Social choice theory provides a framework to aggregate preferences, but was not developped for  the  multidimensional applications typical of AI. 
Leveraging insights from a recently published urn process, this work  introduces  a  preference aggregation strategy that adapts to the user's context and that inherits the good properties of  the maximal lottery, a  Condorcet-consistent  solution concept.
\end{abstract}


\newcommand{\BibTeX}{\rm B\kern-.05em{\sc i\kern-.025em b}\kern-.08em\TeX}

\begin{document}

\pagestyle{fancy}
\fancyhead{}

\maketitle

 \section{Introduction}
The rapid evolution of foundation models has sparked a global race among AI labs, which are fiercely competing to develop the highest-performing models. This is because foundation models underpin a wide range of applications, with game-changing implications for business, scientific advancement, and societal development. The generative artificial intelligence market is expected to grow to \$1.3 trillion over the next 10 years from a market size of just \$40 billion in 2022, according to a new report by Bloomberg Intelligence~\cite{bloomberg}.

One of today's major challenges is alignment: how to ensure that the AI agent behaves in a way that is consistent with human values and preferences, and one key tool to achieve this goal is reinforcement learning with human feedback (RLHF)~\cite{ziegler2019fine,bohm2019better,IntroducingChatGPT2024,NEURIPS2020_1f89885d,NEURIPS2022_b1efde53}, where human annotators provide their preferences to guide the model's training.
The human inputs are then used to fine-tune the model through a supervised learning process, where the model is explicitly trained on labeled examples reflecting these preferences.

Like many other machine learning approaches, RLHF  consolidates indistinguishable users into a single entity. This approach stems from the supervised learning paradigm, where the goal is to predict an output based on an input, and the training data includes a noisy version of this input-output relationship. However, the limitations of this method in the context of preference aggregation are well-documented in~\cite{siththaranjan2023distributional}.

More broadly, multiple studies have highlighted the limitations of current alignment approaches~\cite{Aroyo_Welty_2015,ji2023ai,xu2023rlhf,chakraborty2024maxmin,noothigattuAxiomsLearningPairwise2020}, in particular the incomplete modeling of diverse human preferences, and the challenge to reconcile diverse and possibly antagonistic human preferences into one AI agent. It also appears that modeling human preferences improve performance and scalability~\cite{askellGeneralLanguageAssistant2021a}. This prompted some in the AI research community to explore social choice theory, both for the evaluation and for the training of AI agents~\cite{conitzerSocialChoiceShould2024,ge2024axioms,maura2025jackpot,lanctot2024softcondorcetoptimizationranking,lanctotEvaluatingAgentsUsing2023,dai2024mapping}. 
This shift arises from the realization that social choice theory offers well-established frameworks for aggregating diverse human preferences.

An important solution concept in social choice theory is that of Condorcet winner~\cite{brandt2016handbook}.
A Condorcet winner is a candidate in an election who would win a majority of votes in a head-to-head comparison against each of the other candidates.
A voting method is said to be Condorcet consistent if it always selects the Condorcet winner whenever one exists. Similar notions have been rediscovered in different context by different community (because they hide the ubiquituous concept of zero-sum game Nash equilibrium), for example, people interested in Dueling bandits might have heard of the von Neumann winner~\cite{pmlr-v40-Dudik15}, while other might have heard of the randomized Condorcet winner. 

\begin{figure}
    \centering
    \includegraphics[width=1.\linewidth]{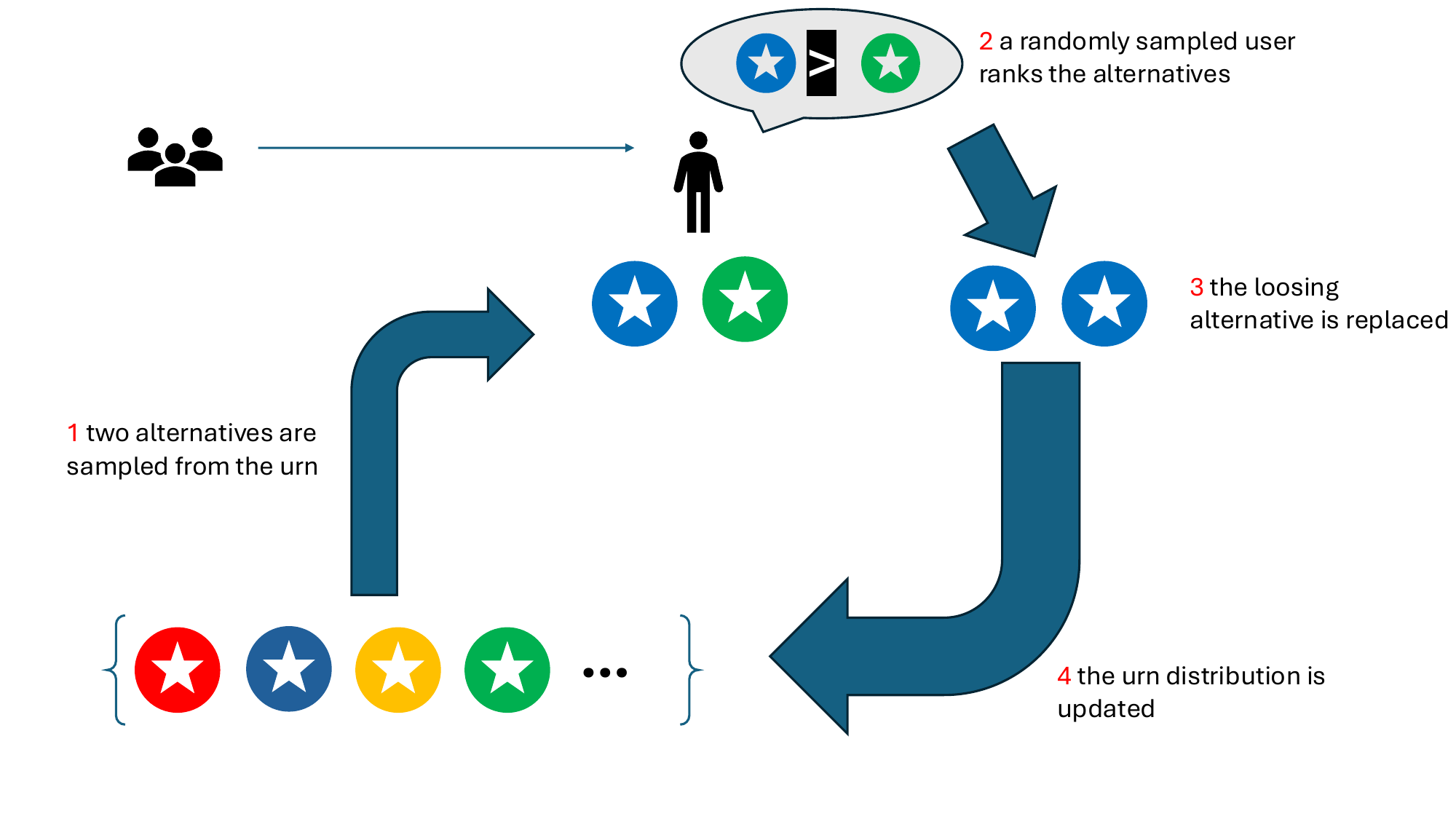}
    \caption{Graphical representation of the urn process introduced by~\cite{brandlNaturalAdaptiveProcess2024}, for simplicity we omit the mutation rate in this representation. 
    Iteratively, (1) two alternatives are sampled from the urn, then (2) a randomly sampled user express their preference of the two options, (3) the ball of the least  prefered option is then replaced in the urn by a ball for the  prefered option, which (4) changes the states of the urn. 
    }
    \Description[Graphical representation of the algorithm introduced by~\cite{brandlNaturalAdaptiveProcess2024} ]{The procedure consists of repeatedly selecting two alternatives at random from the urn and having a randomly chosen participant express their preference between them. The ball representing the losing alternative is then replaced in the urn by another ball representing the winning alternative.}
    \label{fig:urn_process}
\end{figure}

In~\cite{brandlNaturalAdaptiveProcess2024}, the authors propose a very simple balls-and-urn process that converges to a maximal lottery, a solution concept known as Condorcet consistent. 
The process involves a single urn containing balls, where each ball represents an alternative. Initially, the urn is filled with multiple balls for each alternative. The procedure consists of repeatedly selecting two alternatives at random from the urn and having a randomly chosen user (or voter, in the paper's context) express their preference between them. The ball representing the losing alternative is then replaced in the urn by another ball representing the winning alternative. The general idea is picture in Figure~\ref{fig:urn_process}.

We use the main idea of~\cite{brandlNaturalAdaptiveProcess2024} to introduce a novel algorithm for AI alignment.
Our innovation was prompted by the 
  parallels between this urn and ball mechanism and the iterative feedback mechanisms used to fine-tuning large language models (RLHF).
  Similar mechanisms can also be seen in industrial recommender systems (e.g. music recommendation) to warm start the system to a given user tastes. 
We show that, using some function approximation tricks~\cite{hennesNeuralReplicatorDynamics2020,d2024solving}, the urn process idea can be ported to the world of generative AI.
By bridging the fields of voting theory and generative AI, this work lays the groundwork for new approaches to AI fine-tuning,  one of the most pressing challenges in the field.

Our goal in this research is to show that processes similar to the one~\cite{brandlNaturalAdaptiveProcess2024} could be used in an online context to identify maximal lotteries.  Because recommender systems~\cite{boutilierModelingRecommenderEcosystems2023a} and LLMs share many similarities,   this work can be applied to both domain. In fact, the formalization we propose in Section~\ref{subsec:mathematical-model} does not distinguish the two.

\subsection{Contribution}
Our contributions are as follows: 
\begin{enumerate}
    \item we propose a novel optimization criterion for preference aggregation, that we name Adaptive Preference Aggregation (APA). \textit{Adaptive} mirrors the title used to introduce the urn process in~\cite{brandlNaturalAdaptiveProcess2024}
    \item we develop an algorithm to optimize this criterion within a given hypothesis class: Adaptive Preference Aggregation (APA), the online nature of the algorithm makes it very relevant for RLHF, as it will tend to focus the queries to annotators on important comparisons, we show the soundness of the method on generated datasets. 
\end{enumerate} 

We introduce the problem we want to address in Section~\ref{seq:incontextpreferenceaggregation}. We then present the main ingredient of Adaptive Preference Aggregation in Section~\ref{sec:algorihm}.
APA is presented in Section~\ref{sec:algorihm}, and experiments results reported in~\ref{sec:experiment}.
The last section discusses limits and possible follow-up of this work.  

\subsection{Concurent work}
We are aware of contemporary work that points toward the direction of using maximal lotteries, in particular~\cite{maura2025jackpot}, but their methodology is different and complement our claim. 
Before that~\cite{munos2024nashlearninghumanfeedback,swamyMinimaximalistApproachReinforcement2024} introduce a zero-sum game that corresponds in spirit to the game that defines the maximal lottery (but without making the connection with maximal lotteries). 
This connection was made in~\cite{wang2023rlhf}, where the authors analyze RLHF under the lens of reinforcement learning theory. Our perspective is quite novel in that it originate directly from the social choice theory algorithm from Figure~\ref{fig:urn_process}. 

\section{preference aggregation at the user level}
\label{seq:incontextpreferenceaggregation}
In this section, we first present the environment we envision (Section~\ref{subsec:mathematical-model}). We then justify our social choice theory inspired learning objective (Section~\ref{section:learning-objective}).
Last, we discuss the limits of reward based preference aggregation (Section~\ref{sec:scoring-method}).
\begin{figure}
 \centering
\includegraphics[width=.8\linewidth]{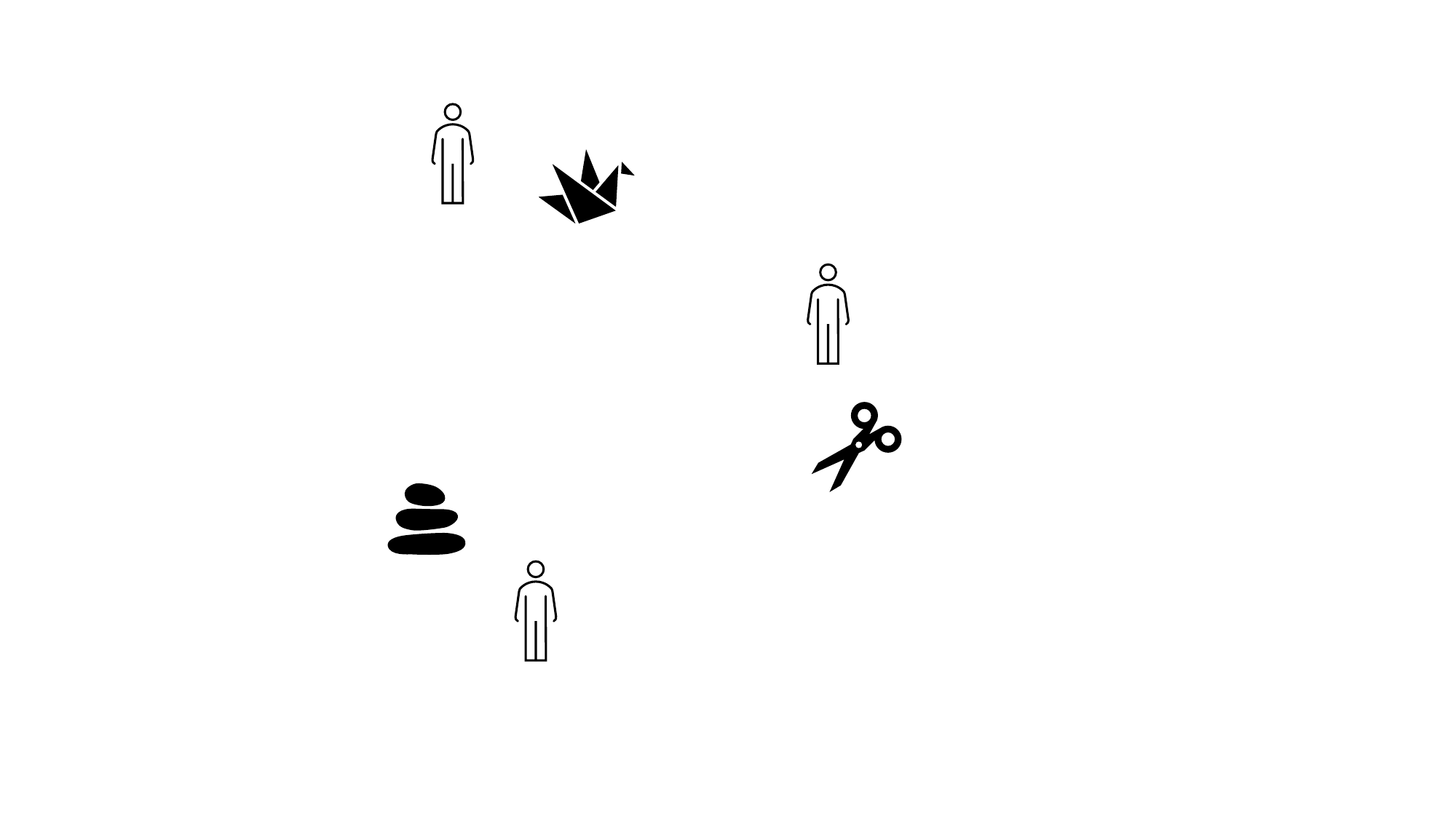}
\caption{Illustration of the mathematical model of Section~\ref{subsec:mathematical-model}.
$\mathcal{A}$ is embodied by the three action of rock-paper-scissors. The users, in white, prefer alternatives that are closer to them. The criterion from \eqref{eq:criterion} induces a non-transitive structure similar to the classical zero-sum game rock-paper-scissors. It is notable that such structure is not well captured by reward based approaches, as explained in Section~\ref{sec:scoring-method}.  }
\Description{A rock paper scissors situation can occurs in the plan}
\label{fig:RPS}
\end{figure}

\subsection{Mathematical model}
\label{subsec:mathematical-model}
We consider a finite set of alternatives $\mathcal{A}$ over  which users from a set $\mathcal{U}$ have preferences. There exists a probability distribution $\mathbb{P}$ over the set of users $\mathcal{U}$, so that $(\mathcal{U},\mathbb{P})$ is a probability space. The preference of each user $u\in \mathcal{U}$ is encoded with a partial order over $\mathcal{A}$ that we denote by $\leq_{u}$: for $a_1$ and $a_2$ in $\mathcal{A}$, $a_1\leq_u a_2$ simply means that the user $u$ prefers alternative $a_2$ over $a_1$.

Maintaining a comprehensive representation of each user's preference relation is infeasible (we do not have access to everything, and we need to encode this in a computer), which motivates mapping users into an embedding space $\mathcal{E}$ using the function $\phi:\mathcal{U}\to \mathcal{E}$. 
When we are given the task of selecting an alternative in $\mathcal{A}$ for  a user $u_0$ -- who we only know by her embedding representation $\phi(u_0)$ -- we would ideally  select  a maximal element of $(\mathcal{A},\leq_{u_0})$. However, (1) we do not know $u_0$, (2) a maximal element in $(\mathcal{A},\leq_{\phi(u_0)})$ does not make any sense (yet). 
This is where   preference aggregation kicks in, see next Section~\ref{section:learning-objective}.
The representation of $u$, $\phi(u)$, may contains features known about the user, past interactions, demographic data, information about their preference... we abstract away the specifics here to focus on how the urn process can be ported to this general setting.

We now highlight an important modeling nuance. In discrete choice theory, an agent's decision is typically modeled as stochastic. The randomness is often attributed to exogenous, unobservable random shocks influencing the decision-making process. 
In contrast, we assume each agent has a deterministic preference relation. The observed stochasticity arises not from inherent randomness in the agent's decision-making, but from the inability to distinguish between agents with distinct preferences using the available data. This perspective shifts the source of uncertainty from exogenous shocks to the limitations of observational data. It also clarifies how preference data can be  interpreted as non-transitive.

\subsection{Learning objective}
\label{section:learning-objective}
Consider two possible alternative answers  to a query, \(A\) and \(B\). It is natural to say that \(A\) is  better than \(B\) if a majority of users prefers \(A\) over \(B\). Extending this idea, a "maximal" alternatives is one that is preferred by a majority of users over \textit{any} other possible alternatives. This maximal  alternatives is commonly referred to as the \textbf{Condorcet winner}, or when we allow to randomize over alternatives, as the \textbf{maximal lottery}. Maximal lotteries are known to have several interesting properties that we do not discuss in this paper~\cite{fishburn1984probabilistic,brandlNaturalAdaptiveProcess2024,brandtFishburnsMaximalLotteries}

To be more formal: given some users $(u_1,\ldots u_n)\in \mathcal{U}$ sampled from $\mathbb{P}$, we consider the task of identifying, using $(u_1,\ldots u_n)$,
a policy $\pi$ from $\mathcal{E}$ to $\mathcal{A}$ that maximizes
\begin{align}
\label{eq:criterion}
  \min_{\pi':\mathcal{E}\to \mathcal{A}}  \mathbb{P}\left(\pi(\phi(u))\geqslant_u \pi'(\phi(u))\right).
\end{align}
This maxmin formulation is reminiscent of the game formulations discussed in ~\cite{munos2024nashlearninghumanfeedback,swamyMinimaximalistApproachReinforcement2024,maura2025jackpot}.

To make this formulation more vivid, 
in the context of the famous Chatbot arena~\cite{chiang2024chatbot}, $\pi$ would be our LLM, $\pi'$ would be any competitor's model, and our goal in expression~\eqref{eq:criterion} corresponds to maximizing our score against our strongest contender. It is important to precise here that the notion of strongest contender is a priori specific to \textit{our} model as preference relations can be intransitive, and we might have in practice a Rock-Paper-Scissor like contest, this perspective connects with the literature on evaluating agents~\cite{omidshafiei2019alpha,balduzziReevaluatingEvaluation2018,lanctotEvaluatingAgentsUsing2023}. 
A pure solution to this problem is called a Condorcet winner, but it might not exist. However, existence is guaranteed when we allow $\pi$ to be stochastic. In this case a solution that maximizes~\eqref{eq:criterion} is called a randomized Condorcet winner or a maximal lottery.

\subsection{Scoring methods for preference aggregation}
\label{sec:scoring-method}
We next present two approaches, one from the world of recommender systems (RecSys), and one from the world of LLM, that learn to rank based on a scoring method, and then present some of  their limits.

\subsubsection{Bayesian Personalized Ranking from Implicit Feedback (BPR)}

A popular recommender system technique called Bayesian Personalized Ranking~\cite{rendleBPRBayesianPersonalized} (BPR) shares some similarity with RLHF. We reinterpret the justification of BPR within our setting, where uncertainty arises not from the user but from the user embedding. 

This reinterpretation leads to a presentation that differs from the original work.
BPR is based on using implicit rankings derived from dataset interactions to predict the rankings of new alternatives for a specific user. In~\cite{rendleBPRBayesianPersonalized}, implicit preferences refer to the relative rankings inferred from user behavior. For example, in a recommender system, a clicked item is assumed to be preferred over an unclicked one. In our setting, feedback is explicitly provided, but the structure of BPR remains relevant due to its connection with the Elo scoring system.

The central structuring assumption in BPR, relevant to our analysis, is that for any user \(u \in \mathcal{U}\) and two alternatives \(a_1\) and \(a_2\), the preference probability is given by
\begin{align}
  \mathbb{P}(a_1 \leq_u a_2 \mid \phi(u)) = \sigma( x_{\phi(u),a_1} - x_{\phi(u),a_2}),
\end{align}
where \(\sigma\) denotes the sigmoid function.
The method assumes a hypothesis class \(\Theta\) for \(x_{u,a}\), expressed as \(x_{u,a} = f(\theta, \phi(u), a)\) with \(\theta \in \Theta\). The solution involves maximizing the likelihood, augmented by a Gaussian prior to enable quadratic regularization.
\textbf{In the context of  Recommender Systems, APA can be seen as an alternative to BPR that will handle in a principle manner non-transitivity of the preference data. }

\subsubsection{RLHF}
Reinforcement Learning from Human Feedback (RLHF) is a technique used to train machine learning models, particularly in the context of aligning LLM behavior with human values and preferences. This method involves a multi-step process where a model is initially trained using standard supervised learning techniques on a dataset. Following this, human feedback is incorporated to fine-tune the model's behavior. This feedback is typically provided by human evaluators who rate or rank the model's outputs based on desired criteria, such as helpfulness or harmlessness.

The reward function plays a crucial role in RLHF. It is constructed based on the human feedback to quantify the desirability of the model's actions or outputs. The model is then optimized using reinforcement learning algorithms to maximize this reward function. Essentially, the reward function serves as a guide, encouraging the model to produce outputs that align more closely with human preferences. 

\subsection{Limits of score based methods}
As mentionned in the introduction, there is a rapidly growing literature
\cite{swamyMinimaximalistApproachReinforcement2024,chakrabortyMaxMinRLHFEquitableAlignment2024,conitzerSocialChoiceShould2024,geAxiomsAIAlignment2024a,xuRLHFIIAPerverse2024,lambert2024reinforcement,bai2022traininghelpfulharmlessassistant,NEURIPS2023_a85b405e,mishraAIAlignmentSocial2023} that always goes back to the limits of score based methods. 
In our context, it seems sufficient to observe that they cannot account for intrensitive preferences such as the situation described in Figure~\ref{fig:RPS}.

\section{Algorithms}
\label{sec:algorihm}

\subsection{Urn process}
In a recent paper,~\cite{brandlNaturalAdaptiveProcess2024} identifies  an algorithm that converges to
the maximal lottery (see Figure~\ref{fig:urn_process} and Algorithm~\ref{alg:color_assignment}). The story goes as follows. First, assign one color to each alternative; take a number $N$ of balls;
colors the balls at random and place them in an urn. Then the process starts. Take two balls and a user at random;
ask the user for their preferred ball over the two balls; change the color of the loosing ball to match the preferred
color; replace the two balls in the urn. Then, repeat (there is also a small probability-- the \textit{mutation rate}-- of relabelling a ball at random).

Then, (1) the distribution of winning balls converges to the maximal lottery under some technical assumptions, (2) the average proportions in the urn converges to the maximal lottery under some technical assumptions.

\begin{algorithm}
\caption{Urn process (\cite{brandlNaturalAdaptiveProcess2024})}
\label{alg:color_assignment}
\begin{algorithmic}[1]
\STATE Fill an array $\mathrm{urn}$ of size $N$ with random elements from $\mathcal{A}$
\WHILE{not converged}
    \STATE Sample $i,j\sim \text{Unif}[1,N]$
    \STATE  Sample $u\sim \mathbb{P}$ at random
    \IF{$\mathrm{urn}_i<_u\mathrm{urn}_j$}
\STATE    $\mathrm{urn}_i\leftarrow \mathrm{urn}_j$
    \ELSE
       \STATE $\mathrm{urn}_j\leftarrow \mathrm{urn}_i$
        \ENDIF
    \STATE Sample $m\sim \text{Bernouilli}(\epsilon)$
    \IF{$m=1$}
   \STATE Sample $i\sim \text{Unif}[1,N]$
   \STATE Sample $a\sim\text{Unif}(\mathcal{A})$
   \STATE $\mathrm{urn}_i\leftarrow a$
    \ENDIF
\ENDWHILE
\end{algorithmic}
\end{algorithm}
\subsection{Connection with Replicator Dynamics}
The discrete urn process described in~\cite{brandlNaturalAdaptiveProcess2024} can be viewed as a discrete, stochastic approximation to the continuous, deterministic \textit{replicator dynamics}~\cite{taylor1978evolutionary,schuster1983replicator,hofbauer1998evolutionary} found in \textit{evolutionary game theory}. The replicator equation models changes in the frequencies of different species over time based on their fitness. In the urn process, the balls represent individuals of different species, and the replacement rule based on pairwise comparisons determines their fitness. This can be interpreted as a learning process for equilibrium play in a symmetric two-player zero-sum game, where the maximal lottery corresponds to an equilibrium strategy.

\subsection{Function approximation}

Replicator dynamics models the evolution of strategy distributions over time in a population, where strategies with higher payoffs increase in frequency. 
However, this typically involves tabular representations, limiting its scalability to complex domains where function approximation is crucial (e.g., using neural networks).

The paper~\cite{hennesNeuralReplicatorDynamics2020} introduces Neural Replicator Dynamics (NeuRD), a method for adapting the continuous-time replicator dynamics equation  to work with function approximation.
NeuRD addresses the tabular limitation by parameterizing the logits of a softmax policy using a function approximator, such as a neural network. The key trick is to treat the discrete-time updates from replicator dynamics as \textbf{target values for the parameterized logits}.

Instead of directly updating the strategy distribution, NeuRD updates the parameters of the function approximator to minimize the distance between the current logits and the target logits derived from the replicator dynamics equation.
This approach leverages the power of function approximation to generalize across different states and actions, making it suitable for large-scale problems.

This idea of using function approximation appear elsewhere in the literature, for example it is used in~\cite{d2024solving} to solve variational inequalities. In this context, Equation~\eqref{eq:surrogate} is refered to as the \textit{surrogate loss}.

\section{Adaptive Preference Aggregation}
\label{sec:APA}

The Adaptive Preference Aggregation algorithm (APA) (see Algorithm~\ref{alg:Online-Algorithm}) learns a mapping from a fixed user embedding $\phi(u)$ to a probability distribution over the finite set of alternatives $\mathcal{A}$, aiming to approximate a maximal lottery for the population of user in the atom $\phi(u)$. 

Operating in an online fashion, it iteratively refines a neural network that emulate an urn. At each step, a user $u\in\mathcal{U}$ is sampled according to the probability $\mathbb{P}$,  followed by sampling two alternatives $(a_1, a_2)$ with a probability  proportional to their presence in the neural urn $f_\theta(\phi(u))$. The user's preference between $a_1$ and $a_2$ is observed, and the neural network's weights $\theta$ are updated to minimize the distance between the current output of the urn and the target  determined by the urn next state of the urn process. The algorithm maintains the weights of the neural network, which define probability distributions over alternatives for each user embedding, approximating the maximal lottery.

\subsection{Function approximation and neural urn}

\begin{algorithm}[tb]
   \caption{Adaptive Preference Aggregation (APA)}
   \label{alg:Online-Algorithm}
\begin{algorithmic}
   \STATE {\bfseries Initialization:} Initialized MLP weights $\theta$, $r$ mutation rate, t = 0 
   \REPEAT
   \STATE Sample $u_t\sim \mathbb{P}$
   \STATE Set $n = f_\theta(\phi(u))$
   \STATE Sample $(a_1,a_2)\in\mathcal{A}$ using $p_t = n/\sum_{a\in\mathcal{A}}(n_a)$
   \STATE Query user $u$ with $(a_1,a_2)$ and set $(a_1^\star,a_2^\star)=(a_1\vee_u a_2,a_1\wedge_u a_2)$.
   \STATE   Set $n_{new} =\max(n +(e_{a_1^\star}-e_{a_2^\star}), 0)$
   \STATE Update $\theta  = \theta - \eta\frac{\partial||f_\theta(\phi(u))-n_{new}||^2}{\partial\theta}$
   \STATE With probability $r$: 
   \STATE \qquad Sample $a_2^\star\sim n/\sum_{a\in\mathcal{A}}(n_a)$ , $a_1^\star\sim \text{Unif}(\mathcal{A})$
      \STATE \qquad   Set $n_{new} =\max(n +(e_{a_1^\star}-e_{a_2^\star}), 0)$
   \STATE \qquad  Update $\theta  = \theta - \eta\frac{\partial||f_\theta(\phi(u))-n_{new}||^2}{\partial\theta}$
   \STATE t=t+1
   \UNTIL
\end{algorithmic}
\end{algorithm}
We simply use a multilayer perceptron
\begin{align}
    f_\theta:y\to n\in \mathbb{R}_+^{|\mathcal{A}|},
\end{align}
 where $n$ represents the urn state, that is the number of balls $N_a$  for each alternative $a\in\mathcal{A}$ given a user of embedding $y=\phi(u)$, and $\theta$ is the network parameter. We use a ReLu for the last layer. 
\footnote{alternatively we could have 
$f_\theta:y\to g$ where $g$ is a generated alternative to be closer to the LLM setting, or add a softmax layer to the perceptron, to be closer to the standard approaches for modeling probability on a discrete set.}
The update should be
\begin{align}
\label{eq:surrogate}
   \min_{\theta} ||f_\theta(\phi(u))-n_{new}||^2,
\end{align}
where 
\begin{align}
    n_{new} =n_{\text{old}} +(e_i-e_j),
\end{align}
with $e_i$ and $e_j$ being the one hot encoded vector that represent the  alternatives $i$ and $j$ that were sampled for user $u$, and $n_{\text{old}}$ is the state of the neural urn when the user is sampled : $ f_\theta(\phi(u))$.

When the embedding are one hot encoded of populations,  Algorithm~\ref{alg:Online-Algorithm} is close to being equivalent to several Algorithm~\ref{alg:color_assignment} run in parallel.

As explained in~\cite{brandlNaturalAdaptiveProcess2024}, we add a small mutation rate. This means that with a small probability, we set
\begin{align}
    n_{new} =n_{\theta_\text{old}} +(e_i-e_j),
\end{align}
where $e_j$ is taken at random from the urn proportionally to $N$, and $e_i$ is sampled uniformly from $\mathcal{A}$.
For the experiments, we used 2 hidden layers with 32 activations units.
To initialize the urn for a given $N$, we did some learning iterations by assigning random values to $N_{new}$ of amplitude proportional to $N$

\subsection{Oscillations}
\begin{figure}
    \begin{subfigure}{0.5\textwidth}
        \includegraphics[width=\linewidth]{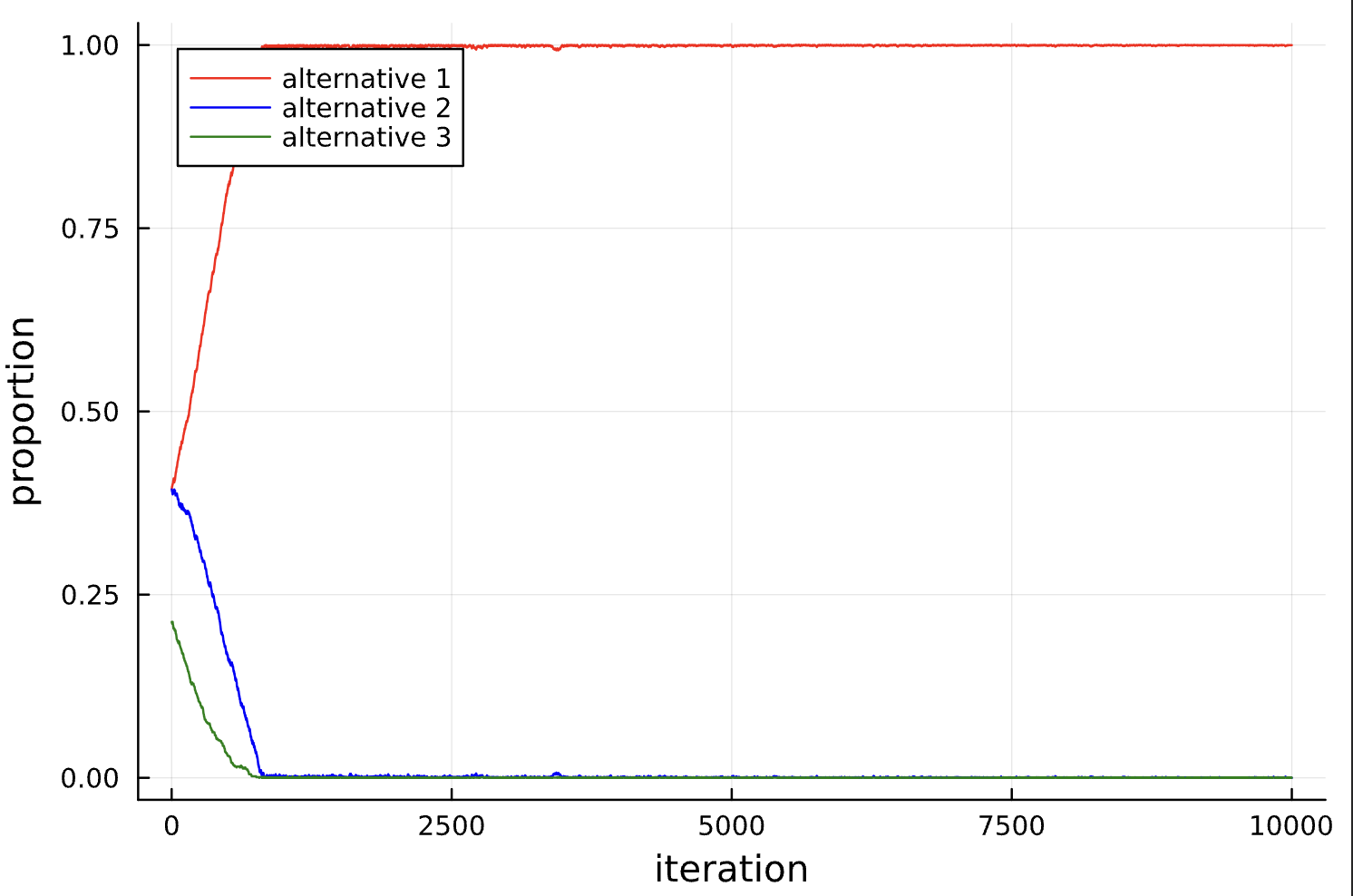}
    \end{subfigure}
    \hfill
    \begin{subfigure}{0.5\textwidth}
        \includegraphics[width=\linewidth]{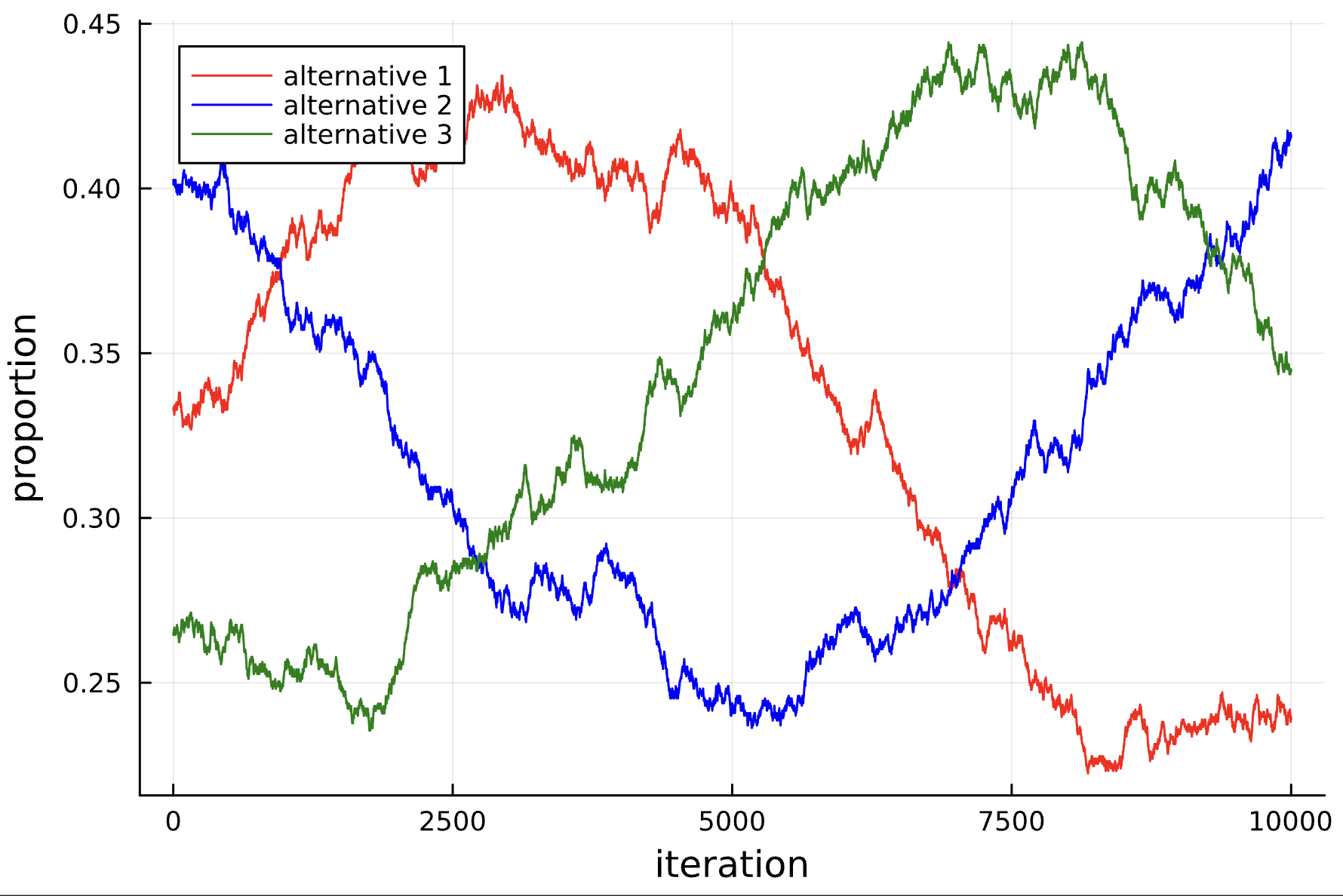}
    \end{subfigure}
   \caption{The algorithm might cycle (top) or not (bottom) depending on the presence of non transitivity in the data. Depending on the downsteam applicartion, this observation might calls for  a distillation step at the end.  }

    \label{fig:need_for_distillation}
\end{figure}

We display in Figure~\ref{fig:need_for_distillation} the components of $N$ in the case where $|\mathcal{A}|=3$ and $\phi(u)$ is a constant. 
We confirm for the case of neural urn the observation made in~\cite{brandlNaturalAdaptiveProcess2024}: the state of the urn might fail to converge, but its time average converges. 
The oscillation on the right-hand side occurs in the presence of non-transitive preferences for the population, such as the one pictured in Figure~\ref{fig:RPS}.
These oscillations can be considered problematic or not, depending on the use case. 
To address them, we propose to average them out using a distillation phase. 
\subsection{Distillation}
The distillation step consists in training a network to predict $p_t$ given $\phi(u_t)$, where $u_t$ was the user sampled at time $t$ in Algorithm~\ref{alg:Online-Algorithm}, and $p_t$ the normalized version of  $f_{\theta_{t}}(\phi(u_t))$. We used a softmax last layer and a cross-entropy loss for this purpose. 
\section{Experiments}
\label{sec:experiment}
In this section, we  report preliminary experiments as well as some practical insights for training APA.

\begin{figure}
\includegraphics[width=0.45\textwidth]{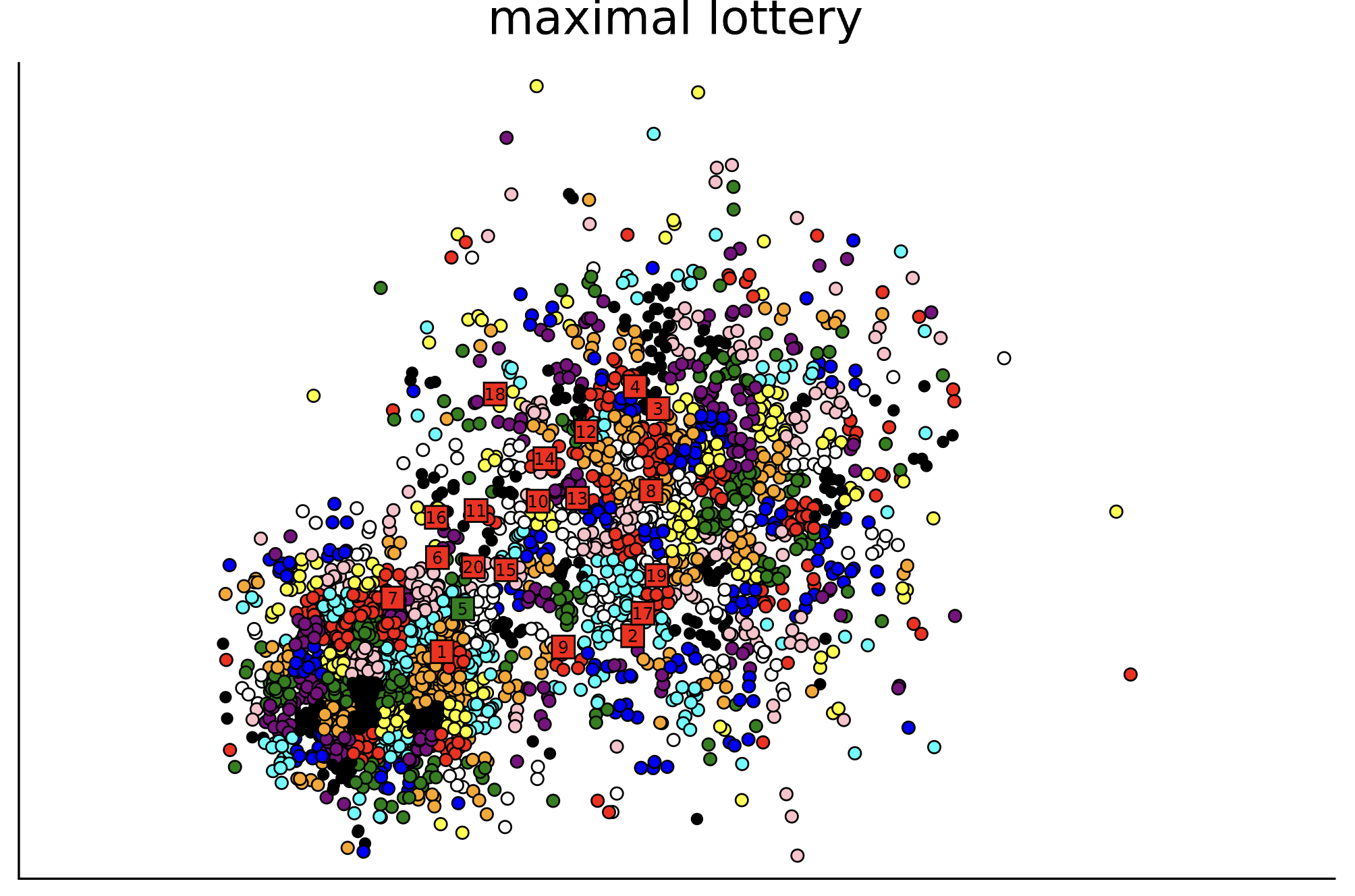}
 \caption{Example of generated data. The squares represent the alternatives, and the points the users. The colors of the points correspond to the atom of the partition induced by the embedding. We represent with a green square the overall Condorcet winner}
 \label{fig:env}
\end{figure}

\begin{figure}
    \centering
\includegraphics[width=0.45\textwidth]{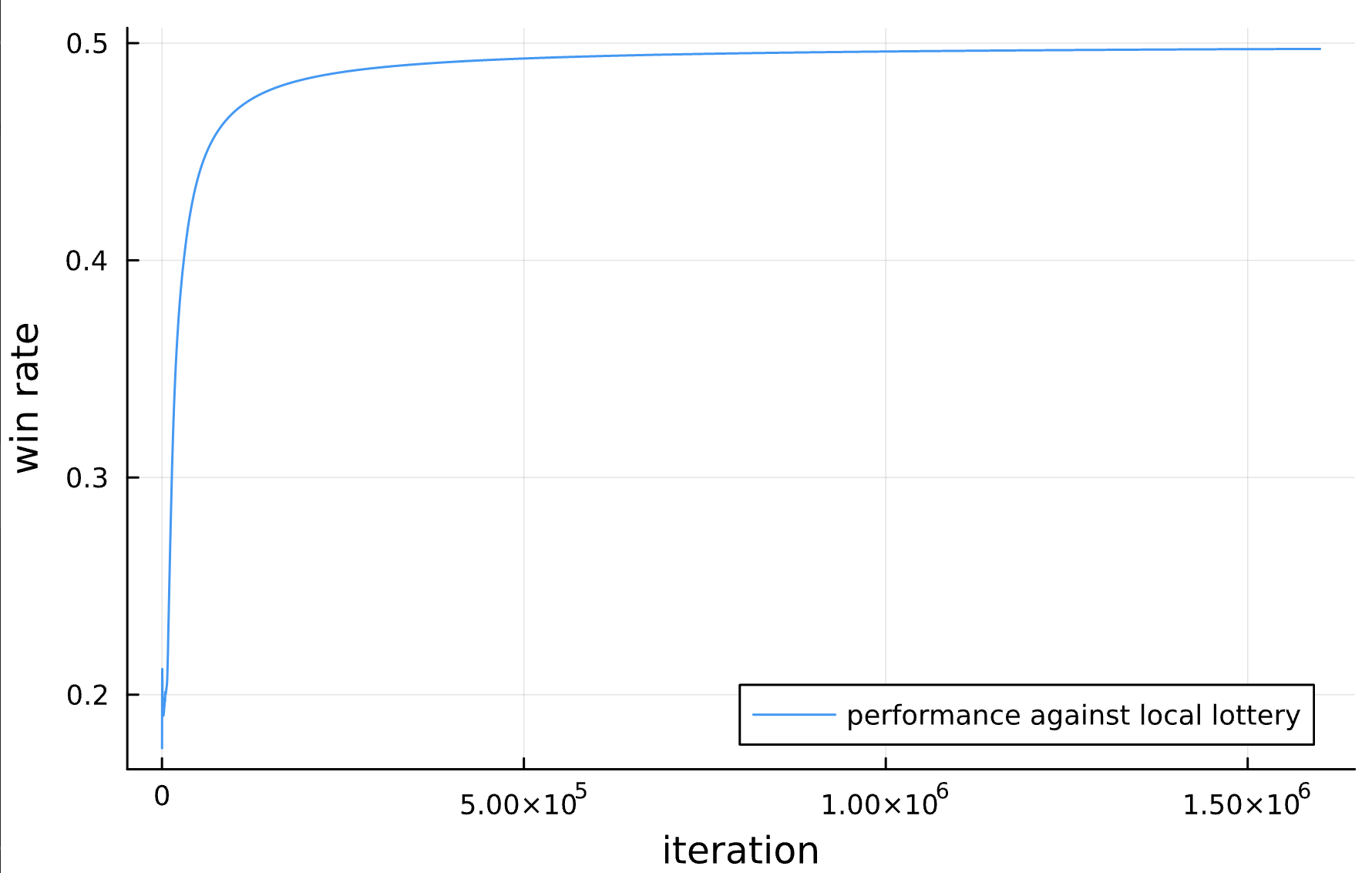}
    \caption{Performance against adaptive maximal lottery during learning (before distillation) }
    \label{fig:perf_before_distillation}
\end{figure}

\subsection{Environment}
 We took $\mathcal{U}=\mathbb{R}^2$, so that the users "true" embedding can be represented in a plan. 
 We then discretize the plan in a  grid to represent the users in the algorithm (this account for the missing information and the algorithm representation of the user, as explained in Section~\ref{subsec:mathematical-model}. Baselines will  suppose access to this discrete grid.
We also embed the element of $\mathcal{A}$ in $\mathbb{R}^2$ and use the preference rule
\begin{align}
    a_1\geqslant_u a_2 \iff ||a_1-u||\leqslant ||a_2-u||.
\end{align}
An example of such environment is displayed in Figure~\ref{fig:env}.

\subsection{Implementation}
The implementation  of the experiment will be provided in the supplementary material, was done in Julia~\cite{Julia-2017}, with the Flux~\cite{Flux.jl-2018,innes:2018}  learning library. 
We use JuMP~\cite{Lubin2023} and GLPK to compute the baselines (local or global maximal lotteries).
For practical reasons, we use a discretization of $\mathbb{R}^2$ to define the user embedding. \textbf{Indeed, this allows us to compare the output of APA with the direct computation of the maximal lotteries on the atoms of the grid using an LP solver. 
}

\subsection{Evaluation methods }
We pit the output of Algorithm~\ref{alg:Online-Algorithm} against maximal lotteries and Borda winners.
The reason to use Borda winner is that, as shown in~\cite{siththaranjan2023distributional}, they correspond to the criterion RLHF is optimizing for. 
Those are computed either at coarse or fine grid: either on all the user available at training time, or conditionned on the users partitions.
\textbf{We can indeed define the equivalents of the \textit{training loss } and \textit{validation loss} by 
 computing, on the  \textit{training set} and a \textit{validation set} of users, a win rate by piting our putative solution against different solutions. }
 Here, we bench against known principled solutions from social choice theory, but in practical setting, one could bench against current state of the art solutions.

\subsection{Evaluation on predistillation data}
\textbf{To follow the full process of APA (with distillation), the evaluation can be done on the data generated during the online phase. } Using the notations of Algorithm~\ref{alg:Online-Algorithm}, we can compare 
$f_{\theta_{t}}(\phi(u_t))$ with $\pi'(\phi(u_t)$ for any opponent policy $\pi'$. Using the familly of maximal lotteries on the atoms of the grid, we consistently obtain plots like Figure~\ref{fig:perf_before_distillation}, with a win rate of $0.5$ which validate our approach.

\subsection{On par with local maximal lotteries}
The reason why we use a coarse grained embedding is that is allows us to compare the output of APA with Social Choice theorical solutions  on the atoms of this coarse grained  embedding. 
We can in particular check that we are on par with the maximal lotteries computed using an LP solver most of the time on
some results  shown in Table~\ref{tab:results}.
\begin{table}[]
    \centering
\begin{tabular}{rr}
\toprule
adaptive maximal lottery & Borda\\
\midrule
0.50 & 0.62\\
0.50 & 0.78 \\
0.50 & 0.76\\
0.50&  0.75\\
0.50 & 0.76 \\
\bottomrule
\end{tabular}
    \caption{Win rate of the distilled neural urn produced by APA against the skyline (adaptive maximal lottery) and the  baselines (Borda)}
    \label{tab:results}
\end{table}

\section{Discussion}
In this work, we clarify a key challenge faced when designing a strategy to aggregate preferences while training systems like LLM or recommender system: the uncertainty surrounding users due to unrevealed information and their encoding within the system. This uncertainty implies that the data may not be sufficient to determine an optimal response to a query. Moreover, the very notion of what is "optimal" becomes unclear.

We propose that one should aim for a sense of Condorcet consistency. We  combine~\cite{brandlNaturalAdaptiveProcess2024} with function approximation to develop an algorithm, APA,  that addresses this gap. We demonstrate, using a non-trivial and visual toy example, that our system can learn the maximal lottery. Further research is needed to understand the conditions under which these observations hold true. Specifically, since the urn process, which inspired our approach, was not originally designed for this type of application, there may be potential improvements in sample complexity and computational efficiency.

From an applied perspective, testing this idea at scale would be an ambitious and valuable follow-up study. Additionally, from a broader viewpoint, the field needs to clarify which properties from social choice theory such systems should possess.

Collaborative filtering, a fundamental concept in recommender systems, often relies on the idea that understanding one aspect of a user's preferences can provide insights into other aspects. We did not fully leverage this perspective in our study, which is a limitation of the present work. Additionally, system performance is frequently evaluated based on a measure of the user's engagement over a relatively short period, which may not align perfectly with the maximality criteria we propose. It remains unclear whether these two perspectives can be reconciled.

From this research, there is a clear need to clarify what generalization means in this context, and to investigate statistical properties such as convergence of estimators for the lottery.

\begin{acks}
The author would like to warmly thank Marc Lanctot for the many insightful conversations around this project. 
\end{acks}

\end{document}